\documentclass{article}

\usepackage{microtype}
\usepackage{graphicx}
\usepackage{xcolor}
\usepackage{amsmath,amssymb,stmaryrd}
\usepackage{subcaption}
\usepackage{fancyvrb}
\usepackage{booktabs} 
\usepackage{hyperref}
\usepackage{xcolor}
\usepackage[framemethod=TikZ]{mdframed}
\usepackage{fvextra} 
\usepackage[T1]{fontenc}
\usepackage[utf8]{inputenc} 
\usepackage{lmodern}
\usepackage{amsmath,amssymb}

\usepackage{listings}
\usepackage[most]{tcolorbox}

\DefineVerbatimEnvironment{modelout}{Verbatim}{
  fontsize=\footnotesize,
  breaklines=true,
  breakanywhere=true,
  breakindent=0pt,
  breaksymbolleft={},
  breaksymbolright={},
  showtabs=false,
  showspaces=false,
  breaksymbolleft={},
  breaksymbolright={}
}

\usepackage[preprint]{icml2026}

\usepackage{amsmath}
\usepackage{amssymb}
\usepackage{mathtools}
\usepackage{amsthm}

\usepackage[capitalize,noabbrev]{cleveref}

\theoremstyle{plain}

\theoremstyle{definition}

\theoremstyle{remark}

\usepackage[textsize=tiny]{todonotes}

\icmltitlerunning{Submission and Formatting Instructions for ICML 2026}

\begin{document}

\twocolumn[
  \icmltitle{Toward IIT-Inspired Consciousness in LLMs: A Reward-Based Learning Framework}



  \icmlsetsymbol{equal}{*}

  \begin{icmlauthorlist}
    \icmlauthor{Hamid Reza Akbari}{yyy}
    \icmlauthor{Mohammad Hossein Sameti}{yyy}
    \icmlauthor{Amir M. Mansourian}{yyy}
    \icmlauthor{Mohammad Hossein Rohban}{yyy}
    \icmlauthor{Hossein Sameti}{yyy}
  \end{icmlauthorlist}

  \icmlaffiliation{yyy}{Sharif University of Technology, Tehran, Iran}

  \icmlcorrespondingauthor{Hamid Reza Akbari}{hamidreza.akbari59@sharif.edu}

  \icmlkeywords{Machine Learning, ICML}

  \vskip 0.3in
]



\printAffiliationsAndNotice{}  

\begin{abstract}
The pursuit of Artificial General Intelligence (AGI) is a central goal in language model development, in which consciousness-like processing could serve as a key facilitator. While current language models are not conscious, they exhibit behaviors analogous to certain aspects of consciousness. This paper investigates the implementation of a leading theory of consciousness—Integrated Information Theory (IIT)—within language models via a reward-based learning paradigm. IIT provides a formal, axiom-based mathematical framework for quantifying consciousness. Drawing inspiration from its core principles, we formulate a novel reward function that quantifies a text's causality, coherence and integration, characteristics associated with conscious processing. Empirically, it is found that optimizing for this IIT-inspired reward leads to more concise text generation. On out of domain tasks, careful tuning achieves up to a \textbf{31\%} reduction in output length while preserving accuracy levels comparable to the base model. In addition to primary task performance, the broader effects of this training methodology on the model's confidence calibration and test-time computational scaling is analyzed. The proposed framework offers significant practical advantages: it is conceptually simple, computationally efficient, requires no external data or auxiliary models, and leverages a general, capability-driven signal rather than task-specific heuristics. Code available at \href{https://github.com/MH-Sameti/LLM_PostTraining}{Github}.
\end{abstract}

\section{Introduction}

\begin{figure}[t]
\centering

\begin{mdframed}[
  linewidth=0.8pt,
  innerleftmargin=6pt,
  innerrightmargin=6pt,
  innertopmargin=6pt,
  innerbottommargin=6pt
]
\setlength{\parindent}{0pt} 
\small

\textbf{Prompt.} \emph{Jan has three times the number of pets as Marcia. Marcia has two more pets than Cindy. If Cindy has four pets, how many total pets do the three have? Let's think step by step and output the final answer after "\#\#\#\#".}\\[0.6em]

\begin{tabular*}{\linewidth}{@{}l@{\extracolsep{\fill}}c r@{}}
\textbf{Baseline} & \textbf{Tokens: 151} & \textbf{Answer: \textcolor{green!50!black}{Correct}} \\
\end{tabular*}

\begin{modelout}
First, we need to determine how many pets Marcia has. Since Cindy has four pets, and Marcia has two more pets than Cindy, Marcia has 4 + 2 = 6 pets. Next, we can determine how many pets Jan has. Jan has three times the number of pets as Marcia, so Jan has 3 * 6 = 18 pets. Now, let's add up the total number of pets the three have. Cindy has 4 pets, Marcia has 6 pets, and Jan has 18 pets. So the total number of pets is 4 + 6 + 18 = 28 pets. Therefore, the three have a total of 28 pets.
#### 28
\end{modelout}

\begin{tabular*}{\linewidth}{@{}l@{\extracolsep{\fill}}c r@{}}
\textbf{[Ours]} & \textbf{Tokens: 42} & \textbf{Answer: \textcolor{green!50!black}{Correct}} \\
\end{tabular*}

\begin{modelout}
Marcia has 4+2 = 6 pets. Jan has 3*6 = 18 pets. Total pets = 18+6+4 = 28.
#### 28
\end{modelout}

\end{mdframed}

\caption{An example showing that a language model inspired by consciousness theory needs fewer reasoning steps to reach a correct final output. Baseline refers to the model post-trained using accuracy as the reward.}
\label{fig:token_example}
\end{figure}

Consciousness is a foundational concept underlying human cognition and societal development~\cite{chalmers2010character, klussman2022importance}, whereby individuals actively identify, analyze, and internalize information about themselves~\cite{morin2011self, eurich2018self, carden2022defining}. Nowadays, language models demonstrate impressive abilities. However, the question of true intelligence goes beyond these achievements. This raises a profound question: Could these advances signal the emergence of machine consciousness comparable to that of humans? Based on existing research, language models have not demonstrated significant capabilities related to self-Consciousness~\cite{chen2025imitation}.

Research on consciousness in \textbf{L}arge \textbf{L}anguage \textbf{M}odels (LLMs) faces several key challenges. The lack of a consensus theory of human consciousness complicates its definition and assessment in LLMs, while existing theoretical frameworks provide limited guidance for LLM-specific research~\cite{butlin2023consciousness}. 

Among these frameworks,\textbf{I}ntegrated \textbf{I}nformation \textbf{T}heory (IIT) is one of the most influential approaches proposed for understanding consciousness~\cite{tononi2004information, albantakis2023integrated}. IIT proposes that the degree of conscious experience corresponds to the extent of integrated information within a system. It offers a mathematical framework to quantify consciousness (via $\Phi$) and qualify its content (via the “conceptual structure” of integrated information). It aims to answer not just where consciousness arises (i.e., in the brain) but why and how it exists, addressing the "hard problem" of consciousness posed~\cite{chalmers1995facing}). However, IIT proponents argue that because AI systems lack the required causal structure, they are almost incapable of generating consciousness~\cite{tononi2004information, findlay2024dissociating, li20250neuroscience}.

This study proposes a framework to enhance consciousness in LLMs using IIT, leveraging reinforcement learning (RL) with a tailored reward function for training. We employ a reward function—namely, the $\Phi$ Reward Function—to quantify properties associated with a conscious system and evaluate it on LLMs. According to IIT, the $\Phi$ value represents the degree of a system’s consciousness with respect to a given state. This study demonstrates that LLMs exhibit limited performance when trained directly using a $\Phi$-based reward function.

To address these limitations, we relax certain constraints involved in computing $\Phi$ and introduce an novel reward function aligned with the cause–effect power property, referred to as the “\textbf{I}ntrinsic \textbf{I}nformation (II) Reward Function”, which is derived from the information postulate. Experimental results demonstrate that LLMs trained with this reward function achieve a substantial reduction in response length (up to 31\%), while maintaining accuracy close to the original performance with appropriate hyperparameter selection.

The main contributions of this paper are summarized as follows:
\begin{itemize}
    
    \item We propose a principled framework for investigating the injection of consciousness-related properties into language models grounded in a theory of consciousness.
    
    \item We propose a novel reward function that reduces response length while preserving model accuracy, offering a simple and computationally efficient approach that relies on a general capability-driven signal without requiring external data or auxiliary models.
    
    \item We evaluate the proposed method in terms of accuracy and efficiency on complex reasoning tasks, comparing against multiple baselines across in-domain settings, out-of-domain generalization, test-time scaling, and confidence calibration.
    
\end{itemize}

\section{Related work}


\subsection{Consciousness Theories}
Consciousness is philosophically puzzling, difficult to define and difficult to study empirically, expert opinions about consciousness—in general, and regarding AI systems—are highly divergent~\cite{butlin2023consciousness}. Nevertheless, multiple theories concerning consciousness have been advanced. Recurrent processing theory posits that recurrent (or feedback) processing within neural circuits is both necessary and sufficient for consciousness~\cite{lamme2000distinct, lamme2010neuroscience}. Global workspace theory(GWT) likens consciousness to a central “stage”where selective information is shared across multiple specialized processors responsible for perception, memory, emotion, and related functions~\cite{baars1993cognitive, dehaene1998neuronal, dehaene2001towards, kalat2014consciousness}. \cite{goldstein2024case} proposes a method to simulate the full GWT process in LLMs via workflow and scheduling without training.~\cite{dehaene2021consciousness}.

\subsection{Targeting LLM Consciousness}
\cite{ding2023survey} demonstrate that GPT-4~\cite{achiam2023gpt} exhibits enhanced self-modeling capabilities by successfully passing a mirror test, although they caution that this does not indicate full consciousness. Similarly, ~\cite{gams2024evaluating} examine ChatGPT through the lens of IIT axioms, observing that it shows advanced levels of information integration and differentiation compared to earlier AI systems, yet remains fundamentally different from human consciousness. \cite{chen2024self} introduce a framework for LLM self-cognition and assess LLMs across four dimensions: comprehension of self-cognition concepts, awareness of their own architecture, expression of self-identity, and the ability to conceal self-cognition from humans. Building on the C0–C1–C2 framework, \cite{chen2025imitation} define LLM self-consciousness and investigate it through benchmark evaluations and analysis of the activation patterns in the models’ internal representations

\begin{figure*}[t]
    \centering
    \includegraphics[width=0.8\textwidth]{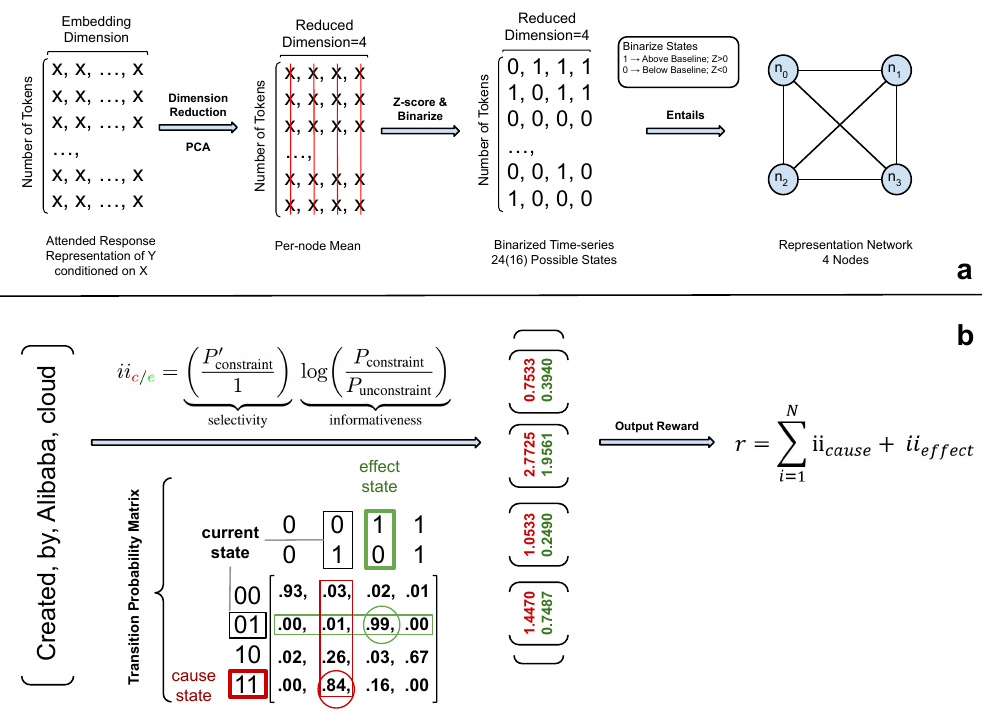}
    \caption{\textbf{IIT-based reward computation}. (a) TPM is constructed from the input sequence x and the generated output y. The representation of y conditioned on x is first refined through an attention-based mechanism. Dimensionality reduction is then applied using Principal Component Analysis (PCA) to project the representation onto a set of units consistent with the IIT framework. The resulting representations are subsequently binarized using mean-based thresholding. (b) Based on the TPM derived from the token sequence, the Intrinsic Information Cause ($ii_{cause}$)  and Intrinsic Information Effect ($ii_{effect}$) are computed for each token. The final reward assigned to the LLM output is obtained by summing the $ii_{cause}$ and $ii_{effect}$ across all tokens.}
    \label{fig:my_figure}
\end{figure*}

\subsection{Datasets and Benchmarks for Consciousness}
\cite{yin2023large} introduce Self-Aware, a dataset of unanswerable questions across five categories paired with answerable counterparts. Building on this, \cite{amayuelas2024knowledge} present Known Unknown Questions (KUQ) and a framework to trace the sources of uncertainty in LLM responses. \cite{li2025knowledge} further formalize LLM knowledge boundaries and provide a comprehensive survey of related research. SA-Bench aims to comprehensively evaluate LLMs’ SA capabilities across three levels: environmental perception, situation comprehension, and future projection~\cite{tang2024towards}. \cite{laine2024me} introduce the SAD benchmark, which applies behavioral tests based on question answering and instruction following across seven task categories.

\subsection{Length Control in LLMs.}

Although reasoning models often achieve strong performance by producing extended chains of thought, this can substantially increase inference latency~\cite{sui2025stop, yeo2025demystifying}. Prior work on length control mitigates this trade-off through techniques such as adjusting positional encodings to enforce a target output length~\cite{butcher2025precise}, incorporating length constraints into the training objective~\cite{jie2023prompt, singhal2023long}, and training on curated datasets annotated with desired reasoning length~\cite{yuan2025following, xia2025tokenskip, biju2025sprint}.

In the context of training objectives, several RL-based post-training methods have been proposed. Some of these approaches control generation length based on a user-selected value~\cite{aggarwal2025l1} or dynamically adjust the number of tokens according to a predefined budget~\cite{han2025token, fang2025thinkless}. On the other hand, some methods define a reward function to encourage a general reduction in the number of generated tokens\cite{arora2025training, luo2025o1, lee2025token}. 

Our proposed framework operates without reliance on curated datasets and eliminates the need for manually specified computational budgets, explicit sequence-length constraints, or predefined task complexity assumptions. Importantly, the observed reduction in output length \textbf{emerges as a structural consequence of increased intrinsic information and system-level consciousness}, rather than being imposed through external constraints. In contrast, most existing approaches achieve length reduction via direct enforcement or heuristic-based length control mechanisms.

\section{Method}

\subsection{Integrated Information Theory}\label{subsec:method_iit}

IIT aims to answer such questions—that is, to offer a scientific account of consciousness, of what it feels like to be you right now.  The IIT method can be summarized as \emph{“consciousness first”}. It starts from the fact that one’s experience exists, right now, and it uses introspection to characterize the properties of experience. The theory then asks how we might account for these properties of experience in physical terms—which means in terms of something we can observe and manipulate “out there.”
The starting point of IIT is the existence of experience. This truth is not the result of an inference; it is rather immediate and confirmed through introspection. The five axioms of IIT express the essential properties of experience, discovered through introspection and reasoning. The five axioms are essential properties of experience because they characterize what experience is—or the nature of experience—and because they are irrefutably true of every conceivable experience, as confirmed through introspection and reflection alone. The axioms can be most succinctly stated as follows :

\textbf{Existence:} Experience exists; there is something.

\textbf{Intrinsicality:} Experience is intrinsic; it exists for itself.

\textbf{Information:} Experience is specific; it is this particular experience.

\textbf{Integration:} Experience is unitary; it is irreducible to separate experiences.

\textbf{Exclusion:} Experience is definite; it is this whole experience.

\textbf{Composition:} Experience is structured; it consists of distinctions and relations forming a phenomenal structure.

According to IIT, A system is any subset of units that can be considered as a candidate substrate of consciousness. A unit is a constituent of a substrate, operationally defined by the grain at which it is observed and manipulated to evaluate its cause–effect power. A substrate of consciousness (also called a complex) is a set of units in a state that satisfies all postulates of IIT. A substrate graph is a way to visually represent the substrate under investigation as a network. It typically consists of the (labeled) units and their causal connections, as nodes and edges of a network respectively. The state of a unit in IIT is binary. A \textbf{T}ransition \textbf{P}robability \textbf{M}atrix (TPM) is a mathematical object that completely describes the dynamics of a substrate of units. More precisely, it is a matrix of numbers that can be used to find the probability that a substrate state will transition from one to another.  

A central issue concerns how the level of consciousness of a given substrate can be determined. The procedure can be understood as six sequential steps, each checking whether a candidate substrate satisfies each of the postulates (0th plus five)\cite{albantakis2023integrated}. A conscious system generates integrated information, measured as $\phi$.

\begin{enumerate}
    \item \textbf{Existence:} Define a substrate model.
    \item \textbf{Intrinsicality:} Select a candidate complex.
    \item \textbf{Information:} Compute intrinsic information.
    \item \textbf{Integration:} Compute integrated information ($\phi_s$).
    \item \textbf{Exclusion:} Identify the main complex.
    \item \textbf{Composition:} Unfold the $\Phi$-structure of the main complex.
\end{enumerate}

\subsection{Problem Setup}

Let $\pi_\theta$ denote an autoregressive LLM policy

\begin{equation}
\pi_\theta(y \mid x) = \prod_i \pi_\theta(y_i \mid x, y_{<i}),
\end{equation}

where $x$ is the input prompt and $y = y_1, \dots, y_{|y|}$ is the output sequence. For clarity, we will suppress the conditioning on x whenever it is clear from the context.

Let $r(y)$ denote a reward function assigning a scalar score to an output sequence. RL seeks to maximize the expected reward

\begin{equation}
\mathbb{E}_{y \sim \pi_\theta}[r(y)].
\end{equation}

The model's consciousness of the generated text is equivalent to the sum of the consciousness of each individual token so

\begin{equation}
r(y) = \sum_{i=1}^{|y|} r(y_i).
\end{equation}

The following sections detail how the consciousness associated with each token is computed.

According to IIT, a system is composed of interconnected units with binary states. The transition probabilities between these states are defined by a TPM. A key question is how a LLM can be utilized to construct such a system—including its units and TPM—from generated texts. 
For this purpose, the token embeddings of the LLM serve as system states. Initially, the output representation is adjusted according to the input representation via an attention mechanism. Due to the high-dimensional nature of LLM embeddings, the resulting state space becomes prohibitively large. Moreover, given the limited number of tokens, dimensionality reduction of the representations is unavoidable. To binarize the system states, the mean value across all representations is computed, with values above the mean thresholded to 1 and those below to 0. The number of system units corresponds to the embedding dimensionality. Since all layer inputs contribute to representation generation within the LLM, the associated units are assumed to be fully connected. This approach is adopted from \citep{li20250neuroscience}. The above steps are illustrated in Figure \ref{fig:my_figure}(a).

To construct the TPM, each consecutive pair of tokens generated by the LLM is interpreted as a state transition from the first token to the second. Consequently, for a sequence of $n$ tokens, $n-1$ transitions are identified and incorporated into the TPM. By normalizing the frequency of these transitions, the resulting matrix is converted into a probabilistic form.

\subsection{Reward Function}
The reward function in this study is derived from two methods—$\Phi$ and II. The phi-based method incorporates all postulates of IIT, while the II approach employs only the information postulate.

\textbf{Phi ($\Phi$)}

As stated at the end of \ref{subsec:method_iit}, the computation of a state-level consciousness measure is performed via a six-step procedure. In the proposed framework, the scalar quantity $\Phi$ is defined as a token-level metric that quantifies the degree of consciousness attributed to each token. A high $\Phi$ indicates that the system’s elements interact in a way that \textbf{cannot be reduced to their individual contributions}, implying a unified conscious state. For computational details, please refer to \cite{ComputingPhi}. The \cite{PyPhi} library was used to compute $\Phi$.

\textbf{Intrinsic Information}

According to the information postulate, a system’s cause–effect state is determined by computing intrinsic information (II) and selecting the cause and effect that maximize its value. II quantifies the system’s specific cause–effect power in its current state. Systems with high II are both \textbf{deterministic} and \textbf{reversible}, exhibiting robust causal properties whereby each state uniquely arises from its preceding state and deterministically transitions to its subsequent state (The consequences of these properties can be observed in the example provided in Figure \ref{fig:token_example}).

\subsection{Why Intrinsic Inforamtion Reward?} 

This study aims to investigate which aspects of consciousness in a language model can be meaningfully quantified under IIT. Based on these quantifiable aspects, the model is trained to enhance its consciousness-related features. Importantly, the presence of such features does not imply that the language model itself is conscious.According to IIT, a conscious system is characterized by two fundamental features:

\begin{enumerate}
   \item It has intrinsic cause–effect power.
   \item The experience is irreducible to a set of separate or independent experiences.
\end{enumerate}

A high $\Phi$ value in a LLM indicates that it exhibits the two aforementioned features. One of the main challenges associated with computing $\Phi$ is its high computational cost, which increases exponentially with the number of units, thereby rendering its calculation difficult and impractical for large-scale systems. The fundamental formulation proposed by IIT is computationally intractable and, in certain cases, lacks a well-defined specification\cite{Barrett20190neuroscience, Moon20190neuroscience}

To address this challenge, the present study proposes a new metric, termed \textbf{intrinsic information}, which has lower computational complexity and focuses specifically on the first feature. The II is computed for each output token of the LLM according to the information postulate. In each computation, the target token is considered as the current state of the system. Consequently, by evaluating the II for each token, a quantitative measure of token-level consciousness is obtained. Subsequently, the corresponding cause state and effect state are identified by selecting those states that maximize the II associated with the current state. Formally, conditioned on the current state, two II measures are computed: $ii_c$, corresponding to the cause state, and $ii_e$, corresponding to the effect state. The II of the current state is then defined as the sum $ii=ii_c + ii_e$. Although, in accordance with \textbf{the Principle of Maximal Existence} in IIT, one may alternatively define the II of the current state as $\max\!\left(ii_c,\, ii_e\right)$, this formulation was also evaluated experimentally. Nevertheless, empirical findings demonstrate that the additive formulation yields more favorable performance during the model training process. The computation process of II is illustrated in Figure \ref{fig:my_figure}(b). 

\textbf{II} is the product of two factors, \textbf{informativeness} and \textbf{selectivity}, which are calculated through operations on the cause TPM and effect TPM as described below

\begin{equation}
ii_{\textcolor{red}{c}/\textcolor{green!50!black}{e}}
=
\underbrace{\left(\frac{P'_{\text{constraint}}}{1}\right)}_{\text{selectivity}}
\;
\underbrace{\log\!\left(\frac{P_{\text{constraint}}}{P_{\text{unconstraint}}}\right)}_{\text{informativeness}}.
\end{equation}

Conceptually, informativeness can be thought of as quantifying causal “raw power,” and selectivity as quantifying “control”. In this setting, due to the absence of a pinned state, the cause TPM and the effect TPM are both identical to the system’s original TPM. \textbf{Constrained probability} is the probability of a specific cause/effect state given that the system is set to its specific current state. \textbf{Unconstrained probability} is the average probability of that cause/effect state if the system is initialized in all possible states. \textbf{Selectivity} characterizes the degree to which a mechanism’s cause–effect power is concentrated over a particular state. On the effect side, selectivity is quantified by the constrained effect state, whereas on the cause side, it is quantified by the backward constrained cause state. For computational details, please refer to \cite{ComputingPhi}.


\begin{figure*}[t]
    \centering
    \centering
    \includegraphics[width=\linewidth]{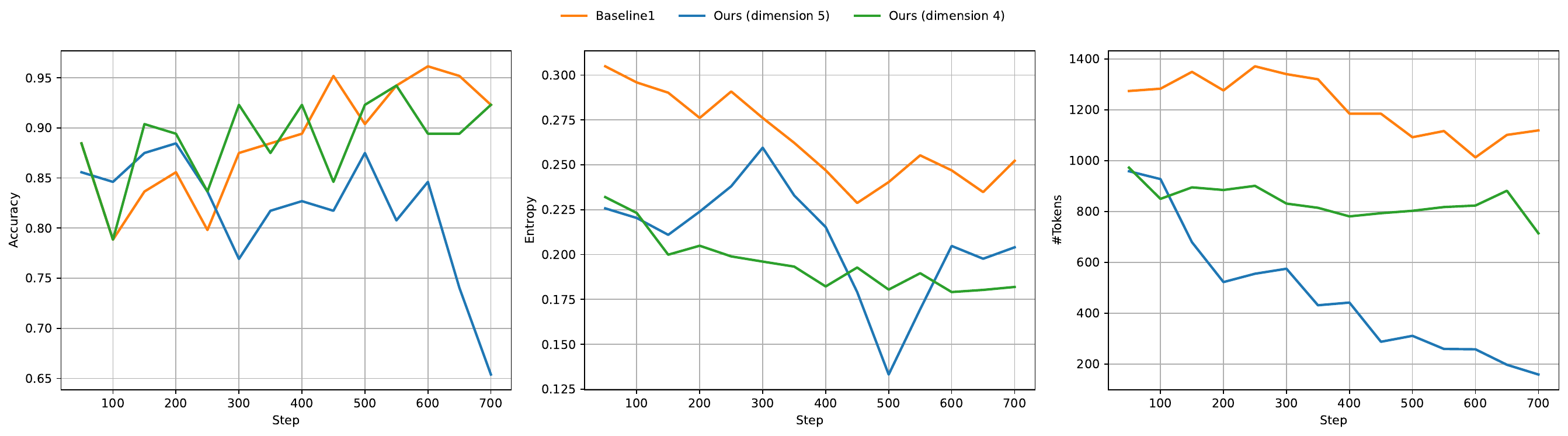}

    \caption{comparison of the proposed method with Baseline 1, in which the model is trained using only an accuracy-based reward. The comparison is conducted on the Open-Thought dataset across different training steps. Evaluation metrics include accuracy,entropy, an the average number of generated tokens.}
    \label{fig:main_results}
\end{figure*}

\section{Experiments}
This section presents preliminary experiments, the experimental setup, training analysis, results, and ablation studies.

\begin{table*}[t]
\centering
\caption{Comparison of accuracy and completion length across benchmarks (Countdown and GPQA evaluated out-of-domain).}
\label{tab:accuracy}
\begin{tabular}{lcccccccc}
\toprule
& \multicolumn{3}{c}{\# Tokens ($\downarrow$)} & \multicolumn{3}{c}{Accuracy ($\uparrow$)} \\
\cmidrule(lr){2-4} \cmidrule(lr){5-7}
Model & Math & Countdown & GPQA & Math & Countdown & GPQA \\
\midrule 
Baseline 0  & 2831.0 & 1188.9 & 5752.7 & 89 & 30 & 17  \\
Baseline 1 & 2929.7 & 2282.8 & 5172.1 & 90 & 37 & \textbf{40} \\
Baseline 2 & 2888.8 & 3123.2 & 6142.0 & \textbf{92} & \textbf{43} & 38 \\
Our Model (Dimension=4) & 2248.0 & 1077.6 & \textbf{3988.3} & 89 & 35 & 23 \\
Our Model (Dimension=5) & \textbf{2058.7} & \textbf{1067.9} & 4845.8 & 89 & 30 & 23 \\
\bottomrule
\end{tabular}
\end{table*}


\subsection{Preliminary Experiments}
To study the behavior of different reward designs, we first conduct a comprehensive set of exploratory experiments using the Qwen-3B-Instruct model \cite{team2024qwen2} on the GSM8K dataset within a post-training framework, employing the GRPO algorithm. Training is performed using Low-Rank Adaptation (LoRA), updating approximately 7\% of the model parameters. Both accuracy-based and consciousness-based rewards are considered: the accuracy reward is binary, assigning a value of 1 for correct answers and 0 otherwise, while the consciousness-based reward ($\Phi$ or II) is defined as a continuous value in the range $[0,1]$ via an exponential mapping.

These experiments evaluate model behavior under a variety of training settings and hyperparameter configurations (details provided in Appendix \ref{sec:appendix A}). Model performance is assessed using accuracy, average response length, and response entropy. The insights obtained from this systematic analysis are then used to inform the design of our main experimental setup, which is presented and discussed in the following section.

\subsection{Experimental Setup}
\textbf{Training data, models, and methods:} DeepSeek-R1-Distill-Qwen-7B~\cite{guo2025deepseek} is adopted as the base model, with OpenThoughts-114k-Math used as the training dataset to construct a stronger reasoning-oriented language model. Based on preliminary experimental results and the characteristics of the selected model and dataset, the TPM construction policy is configured in prompt mode, with reduced dimensionality evaluated at values of 4 and 5. To improve training efficiency, a parameter-efficient fine-tuning (PEFT) strategy is employed, in which only 14\% of the model parameters are updated during training. Configurations with a reduced number of trainable parameters are further evaluated; however, these settings do not achieve the expected performance..

\textbf{Evaluation:} We evaluate the trained model on a diverse suite of mathematical and reasoning benchmarks, including Math-500, Countdown, GPQA-Diamond, covering a wide range of problem difficulties and reasoning requirements~\cite{cao2025toward}. To evaluate the generalization behavior of the trained models with respect to the dataset used during language model training, MATH-500 is considered as the in-domain dataset, while Countdown and GPQA-Diamond are treated as out-of-domain datasets. Meanwhile The trained models are evaluated against the baselines in terms of \textbf{Accuracy} and \textbf{Average Response Length}.

\textbf{Baselines:} To rigorously evaluate our approach, we compare it against three baseline methods, which are detailed below. All models are evaluated using identical inference settings to ensure a fair comparison.

\begin{itemize}
    \item \textbf{Base reasoning model (DeepSeek-R1-Distill-Qwen-7B):} This model is a distillation of the main R1 reasoning model into Qwen-2.5-7B, released by DeepSeek (Baseline 0).
    
    \item \textbf{Base reasoning model with Accuracy Reward:} DeepSeek-R1-Distill-Qwen-7B was trained on the OpenThoughts-114k-Math dataset using the GRPO algorithm with an accuracy reward, thereby isolating the effect of the proposed reward formulation (Baseline 1).
    
    \item \textbf{Base reasoning model with Entropy Minimization:} As reported in \cite{agarwal2025unreasonable}, LLMs can be trained using an entropy-based reward. Their results show that minimizing response entropy improves accuracy. Since their approach, like ours, also reduces response entropy, a comparison of the two methods is well justified (Baseline 2). 
\end{itemize}

\subsection{Training Process Analysis}
An examination of the LLM training process with respect to accuracy, response length reduction, and entropy—evaluated across training iterations and on validation data—provides insight into the stability of the proposed training methodology. Figure \ref{fig:main_results} presents the trajectories of the aforementioned metrics over the course of training. Several key observations can be derived from this figure.

\textbf{First}, the average response length consistently decreases as training progresses. This reduction is notably more pronounced in the model with dimensionality 5 (up to 75\%), corroborating findings from preliminary experiments that suggest an intensified effect with increasing model dimensionality.

\textbf{Second}, the LLM’s accuracy exhibits fluctuations around the Baseline II reference throughout the training process. These fluctuations become more prominent in configurations where a greater reduction in response length is observed, indicating a potential trade-off between response compactness and predictive stability.

\textbf{Finally}, the entropy of the trained models remains lower than that of Baseline II. This reduction in entropy may suggest an enhancement in the strength of causal dependencies underlying the response generation mechanism, leading to more deterministic and structured model outputs

\begin{table*}[t]
\centering
\caption{Test Time Scaling and Confidence Calibration comparison across Math-500, countdown datasets.}
\label{tab:tts_mce_ece}
\begin{tabular}{lcccccc}
\toprule
& \multicolumn{2}{c}{TTS ($\uparrow$)} & \multicolumn{2}{c}{MCE ($\downarrow$)} & \multicolumn{2}{c}{ECE ($\downarrow$)} \\
\cmidrule(lr){2-3} \cmidrule(lr){4-5} \cmidrule(lr){6-7}
Model & Math & Countdown & Math & Countdown & Math & Countdown \\
\midrule
Baseline 0  & 90 & \textbf{58} & 0.4044 & 0.4450 & 0.2138 & 0.2365 \\
Baseline 1 & 91 & 57 & 0.4517 & 0.4517 & 0.2419 & 0.2122 \\
Baseline 2 & 91 & 45 & \textbf{0.3207} & 0.4309 & \textbf{0.1938} & \textbf{0.1328} \\
Our Model (Dimension=4) & 89 & 39 & 0.4134 & 0.4780 & 0.2486 & 0.2288 \\
Our Model (Dimension=5) & \textbf{92} & 33 & 0.4217 & \textbf{0.3683} & 0.2606 & 0.1856 \\
\bottomrule
\end{tabular}
\end{table*}

\subsection{Results}
Table \ref{tab:accuracy} reports the results of the trained models and the baselines across the benchmarks under the considered evaluation metrics. Several observations can be drawn from these results:

\textbf{Response length is reduced across all benchmarks.} The largest reduction of 31\% was observed on GPQA as an out-of-domain task, whereas a 28\% decrease was observed on MATH-500 as an in-domain task.

\textbf{Across all benchmarks, our model demonstrates a notable improvement in accuracy compared to Baseline I.} However, its accuracy is lower than that of Baseline II and Baseline III. This observation may indicate a trade-off between sequence length reduction and accuracy. Nonetheless, the results suggest that combining the accuracy and II rewards improves performance relative to the no-training scenario (Baseline I).

\textbf{On all benchmarks, the response length of Baseline2 increases compared to Baseline1.} This suggests that the increase in response length may stem from training with a reward-based objective. However, Intrinsic Information Reward Function achieves a greater reduction than Baseline1, effectively mitigating the response length inflation commonly associated with reward-based training.

\textbf{While entropy-based training slightly improves accuracy across all benchmarks, it performs poorly in terms of response length reduction.} In particular, the largest increases in response length are observed on Countdown. Consequently, the entropy-based method does not constitute a competitive alternative to the proposed approach.

\subsection{Ablation}
Given the importance of Test-Time Scaling (TTS)~\cite{zhang2025and} and Confidence Calibration~\cite{guo2017calibration} in the development and improvement of LLMs, the proposed method is further analyzed and evaluated from the perspective of alignment with TTS and confidence calibration. Test-Time Scaling is assessed through the implementation of self-consistency. For the evaluation of self-consistency, we generated five responses per data sample and determined the final answer using a majority voting scheme. The temperature parameter was set to 0.7 to control the stochasticity of the model's output. Confidence calibration is evaluated using the Expected Calibration Error (ECE) and Maximum Calibration Error (MCE) metrics. For the purpose of this analysis, the data was discretized into 10 bin intervals. Table \ref{tab:tts_mce_ece} show the results of Test-Time Scaling and Confidence calibration compared to the baselines. 

By comparing the accuracy metric between Tables 1 and 2 across all benchmarks, an improvement in the model’s performance under the self-consistency setting can be observed. Although this improvement is marginal on the MATH-500 benchmark compared to GPQA, the baseline I and II models exhibit a larger accuracy gain on GPQA than our proposed model. In the confidence calibration evaluation, as expected, entropy-based training (Baseline III) demonstrates a greater reduction in both MCE and ECE metrics. Consequently, under this baseline, the predicted probabilities serve as a reliable representative for the correctness of the model’s responses. Although our models reduce the MCE and ECE metrics on certain benchmarks, they fail to achieve comparable performance on others.

\section*{Conclusion}
We presented a framework to enhance feature-consciousness in LLMs via reinforcement learning with an intrinsic information reward based on integrated information theory information postulate. Benchmark evaluations show that models with improved cause-effect power produce more concise responses. While validated on both in-domain and out-of-domain data, extending this approach to more complex tasks—such as very hard mathematical reasoning and code generation—remains a valuable direction for future work.

Additionally, consciousness was quantified through pre- and post-state; given its abstract nature, alternative definitions could provide further insights. 

Given the inherent trade-off between improving predictive accuracy and reducing output length, a systematic analysis of the interaction between these two objectives is required. This analysis should be grounded in training-process variables, such as the in intrinsic information, which can significantly influence optimization dynamics. From an information-theoretic perspective, examining how reward shaping affects this trade-off may yield insights into the emergence of compact yet high-accuracy representations during model training.

Hyperparameter selection poses challenges due to dynamic model behavior during training, such as varying response lengths affecting the data required for transition probability matrix construction. Exploring adaptive or hyperparameter-free strategies represents a promising avenue for further enhancing this framework.

\nocite{langley00}

\bibliography{main}
\bibliographystyle{icml2026}

\newpage
\appendix
\onecolumn
\section{}
\label{sec:appendix A}
\subsection{Hyperparameters}
In the process of determining the number of units and constructing the TPM, the following two hyperparameters are considered:

\textbf{Reduced dimensionality:} This hyperparameter defines the size of the discretized state space used to represent the system. The dimensionality must be compatible with the batch size of the training data to ensure reliable estimation of transition probabilities. Notably, when computing rewards based on the $\Phi$ measure, there is an inherent limitation on the number of dimensions due to combinatorial explosion; given our available computational resources, $\Phi$-based rewards cannot be reliably computed for dimensionalities greater than three. In contrast, rewards derived from II do not impose this dimensionality restriction.

\textbf{Dataset size / TPM construction policy:} Different strategies can be adopted for TPM construction, reflecting the way trajectories or responses are aggregated:

\begin{itemize}
    \item Batch mode: All responses generated by the LLM within a single batch are aggregated to construct a single TPM, representing the collective state transitions of that batch.
    
    \item Prompt mode: Since multiple responses may be generated for a single prompt during training, only responses corresponding to the same prompt are used to construct the TPM for that specific prompt. Consequently, one TPM is generated per prompt, allowing reward computation that is specific to individual prompts.
    
    \item Trajectory mode: A separate TPM is constructed for each individual response, capturing the transition dynamics at the granularity of single trajectories.
\end{itemize}

\textbf{State size:} As discussed in Section 4, the reward assigned to a response is computed as the sum of the rewards for each individual token in the response. In this approach, each state is initially defined as a single token. However, the state size can be extended beyond a single token to include a fixed-width sequence of multiple tokens. For instance, if the state size is set to four tokens, an output sequence of 200 tokens would correspond to 50 states. Increasing the state size allows the model to capture higher-level cause–effect relationships across token sequences, aligning with the principles of transition dynamics in both RL and IIT frameworks. Therefore, the state size is considered in two forms: token-based and chunk-based. In the chunk-based setting, the state size is treated as a hyperparameter.

\subsection{Phi-based Training}
\begin{figure}[h]
    \centering
    \includegraphics[width=0.8\textwidth]{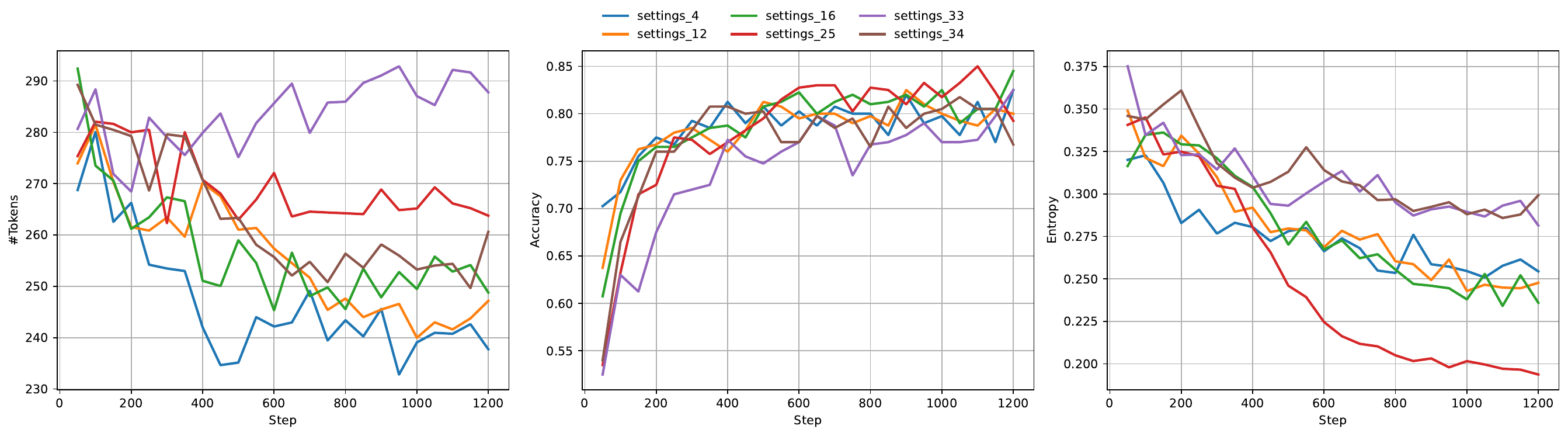}
    \caption{Phi-based reward Training}
    \label{fig:phi_based_reward_training}
\end{figure}

The five different settings used for training the model with the $\Phi$-based reward. The corresponding results, reported according to the evaluation metrics, are shown in Figure \ref{fig:phi_based_reward_training}. In all settings, the reduced state-space dimensionality was set to 3 due to computational resource constraints. Table \ref{tab:hyper_param_phi_settings} summarizes the experimental configurations, detailing the specific hyperparameter values employed in each setting.

The experimental results obtained under various configuration settings can be summarized as follows. \textbf{First}, the majority of the evaluated settings resulted in a reduction in response length; nevertheless, the baseline configuration exhibited the most pronounced decrease. \textbf{Second}, model accuracy remained largely consistent across all settings, with several configurations achieving marginal improvements over the baseline. \textbf{Third}, with respect to entropy reduction, configurations adopting a trajectory-based TPM construction policy substantially outperformed the baseline. Taken together, these results suggest that the Phi reward function does not yield a statistically

\subsection{Intrinsic Information based Training}
Given that the II Reward Function permits the specification of higher-dimensional configurations, a wider spectrum of experimental settings was systematically investigated under this framework. The evaluation outcomes are subsequently reported stratified by the TPM construction policy. Table \ref{tab:hyper_param_ii_settings} summarizes the experimental configurations, detailing the specific hyperparameter values employed in each setting.

\begin{figure}[h]
    \centering
    \includegraphics[width=0.8\textwidth]{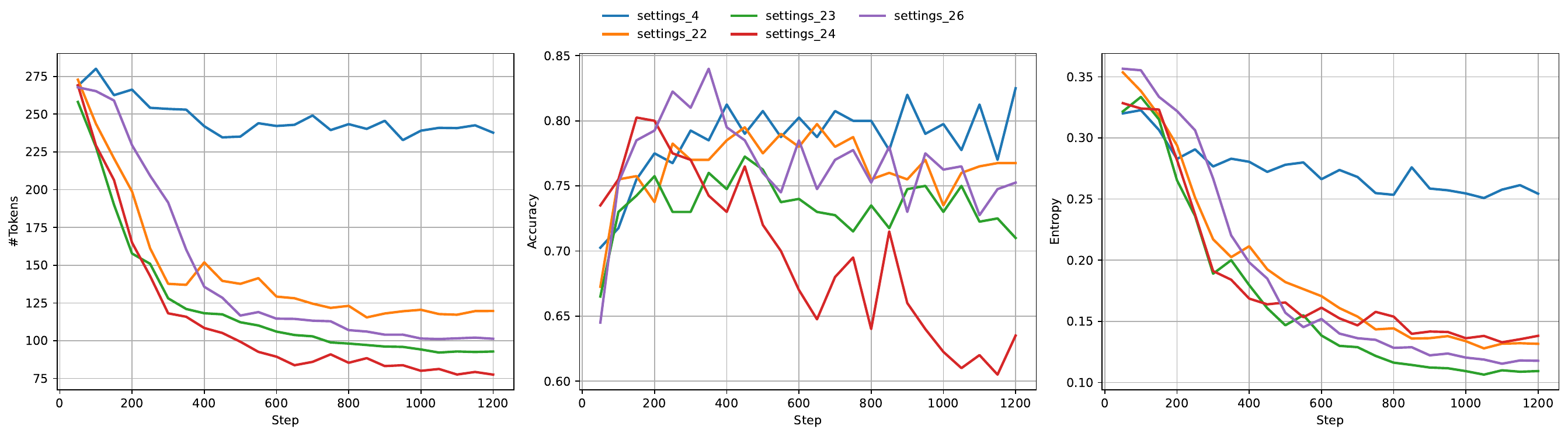}
    \caption{II Reward Training Using Trajectory Mode}
    \label{fig:trajectory_ii_based_reward_training}
\end{figure}

\textbf{Trajectory mode:} As illustrated in Figure \ref{fig:trajectory_ii_based_reward_training}, the experimental results are summarized as follows:

\begin{itemize}
    \item The average sequence length across all experimental settings was significantly reduced by approximately 60\% compared to the baseline.
    
    \item The mean accuracy across all settings was 71\%, corresponding to a 9\% decrease relative to the baseline. Notably, one particular setting exhibited a substantial drop in accuracy due to a mismatch between the TPM size and the number of tokens, which in turn affected the overall average accuracy.
    
    \item Entropy metrics demonstrated a pronounced reduction across all settings, indicating more deterministic or constrained output distributions.
\end{itemize}

In summary, the Trajectory mode led to a marked reduction in both average response length and entropy; however, this was accompanied by a noticeable decline in accuracy. These findings suggest a trade-off between response conciseness and predictive fidelity in this mode.

\begin{figure}[h]
    \centering
    \includegraphics[width=0.8\textwidth]{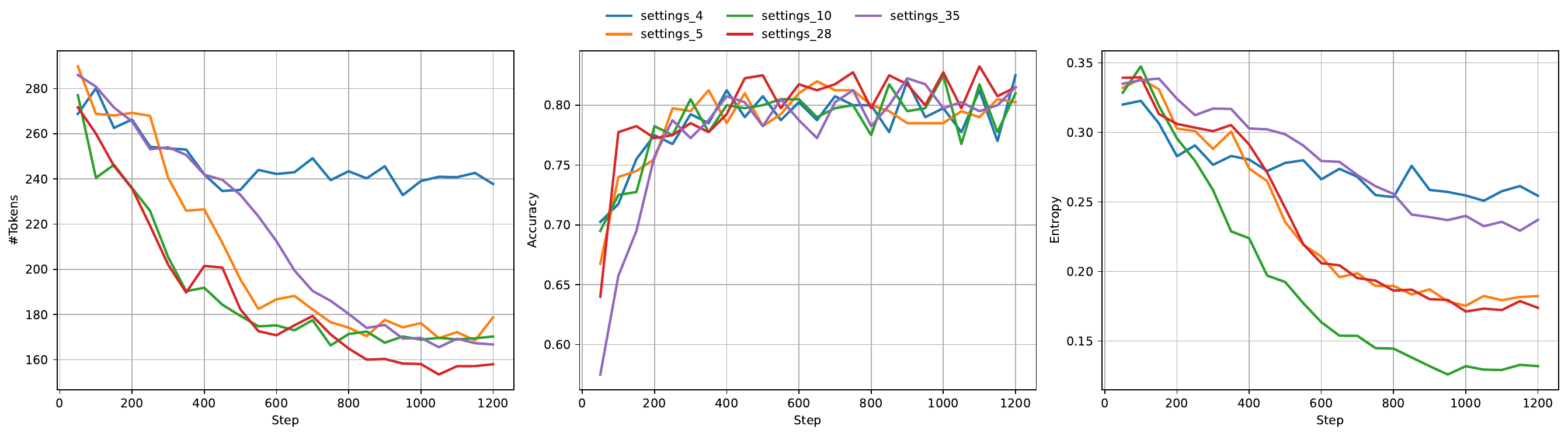}
    \caption{II Reward Training Using Prompt Mode}
    \label{fig:prompt_ii_based_reward_training}
\end{figure}

\textbf{Prompt mode:} Under this mode, four experimental configurations were evaluated in comparison with the baseline. The experimental results for the prompt mode are illustrated in Figure \ref{fig:prompt_ii_based_reward_training}. The results are summarized as follows:

\begin{itemize}
    \item The average output length of the language model was 168 tokens, corresponding to a reduction of over 30\% relative to the baseline. This reduction was consistently observed across all evaluated configurations.
    
    \item The model achieved an average accuracy of approximately 80\%, exhibiting no measurable degradation compared to the baseline.
    
    \item Entropy was reduced across all configurations; however, in certain settings, the magnitude of reduction was less pronounced than that observed under the baseline condition.
\end{itemize}

Overall, the results suggest that the prompt-based TPM construction policy enables a significant reduction in output length while preserving model accuracy. Nonetheless, the reduction in output length is less pronounced than that observed in the trajectory-based mode, underscoring a trade-off between compression efficiency and prompt construction strategy.

\begin{figure}[t]
    \centering
    \includegraphics[width=0.8\textwidth]{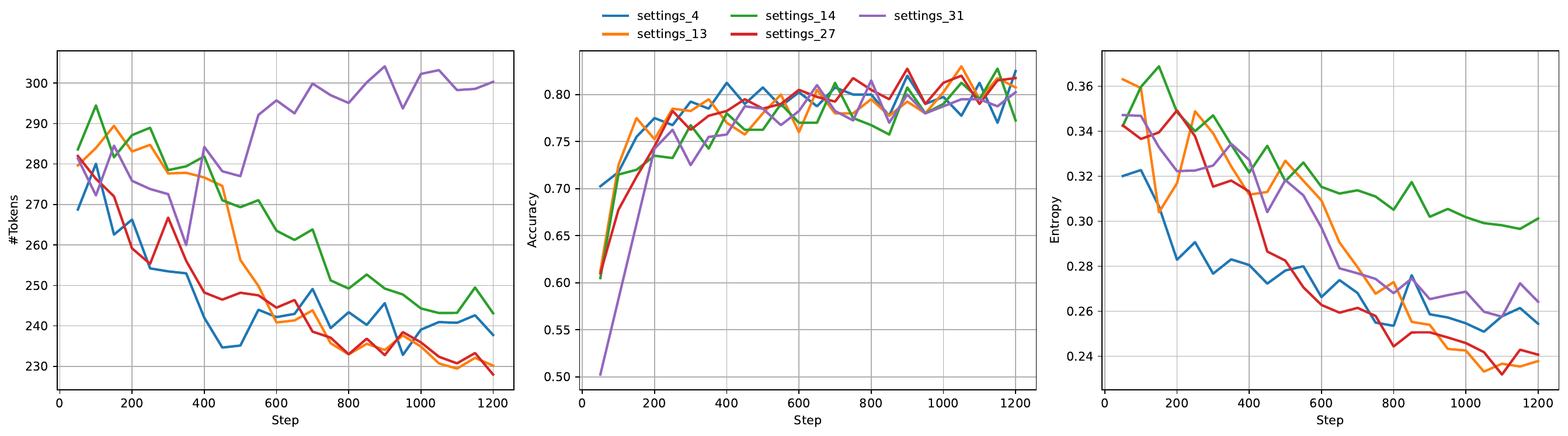}
    \caption{II Reward Training Using Batch Mode}
    \label{fig:batch_ii_based_reward_training}
\end{figure}

\textbf{Batch mode:} As illustrated in Figure \ref{fig:batch_ii_based_reward_training}, the performance of different configurations under the batch-based TPM construction policy can be summarized as follows:

\begin{itemize}
    \item The average output length was 250 tokens, corresponding to a 5\% increase relative to the baseline. Nevertheless, a decreasing trend in output length over the course of training—similar to that observed in the baseline—was evident across most configurations.
    
    \item The model achieved an average accuracy of approximately 80\%, exhibiting no measurable degradation compared to the baseline.
    
    \item Entropy consistently decreased across all configurations during training, following a pattern comparable to the baseline. Notably, in certain settings, the magnitude of entropy reduction exceeded that observed under the baseline condition.
\end{itemize}

Overall, the batch-based TPM construction policy does not yield improvements in output length reduction. However, model accuracy is preserved, and entropy is effectively reduced throughout the training process, indicating stable learning dynamics despite limited gains in response compression.

\begin{table}[h]
\centering
\caption{Summary of Hyperparameter Settings in $\Phi$ Based Training}
\label{tab:hyper_param_phi_settings}
\begin{tabular}{lcccccccc}
\toprule
\cmidrule(lr){2-3} 
Model & TPM Construction Policy & State Size  \\
\midrule
Settings 4 (Baseline) & - & - \\
Settings 12 & prompt & token \\
Settings 16 & batch & token \\
Settings 25 & trajectory & token \\
Settings 33 & batch & chunk \\
Settings 34 & prompt & chunk \\
\bottomrule
\end{tabular}
\end{table}

\begin{table}[h]
\centering
\caption{Summary of Hyperparameter Settings in Intrinsic Information Based Training}
\label{tab:hyper_param_ii_settings}
\begin{tabular}{lcccccccc}
\toprule
\cmidrule(lr){2-4} 
Model & TPM Construction Policy & State Size & Reduced dimensionality  \\
\midrule
Settings 4 (Baseline)  & - & - & - \\
Settings 5 & prompt & token & 3 \\
Settings 10 & prompt & chunk  & 6 \\
Settings 28 & prompt & chunk & 3 \\
Settings 35 & prompt & token & 6 \\
Settings 13 & batch & token & 6 \\
Settings 14 & batch & token  & 3 \\
Settings 27 & batch & chunk & 6 \\
Settings 31 & batch & chunk & 3 \\
Settings 22 & trajectory & token & 3 \\
Settings 23 & trajectory & chunk  & 6 \\
Settings 24 & trajectory & token & 6 \\
Settings 26 & trajectory & chunk & 3 \\
\bottomrule
\end{tabular}
\end{table}

\subsection{Analysis of Experimental Results}
Based on the experimental results, several observations merit discussion.

\textbf{First,} the proposed II Reward consistently outperforms the Phi reward function across the evaluated settings, indicating its stronger capacity to capture informative structural properties of generated sequences.

\textbf{Second,} within the GRPO framework, the final reward signal used for training the LLM is defined as the sum of the “accuracy reward” and the “II Reward”. While the accuracy reward explicitly incentivizes the model toward producing the correct final answer, the “II Reward” operates differently. Specifically, a high II Reward indicates that each generated token is highly constrained by both its preceding and succeeding tokens, thereby encouraging the emergence of stronger cause–effect dependencies within the generated text. This behavior provides a plausible explanation for the observed reduction in response length. Intuitively, when the model exhibits increased consciousness of its current state—characterized by an understanding of the most likely causal predecessor (cause state) and successor (effect state)—it tends to favor more compact generations with reduced redundancy.

\textbf{Third,} increased consciousness of preceding and succeeding states does not necessarily align with improvements in task accuracy. The absence of a substantial accuracy gain suggests that the level of structural consciousness induced by maximizing the II Reward alone may be insufficient to improve accuracy. In some configurations, over-optimization of this reward even leads to a degradation in accuracy, highlighting a potential trade-off between response length and task performance.

\textbf{Fourth,} an additional noteworthy observation is the reduction in the entropy of the generated texts. This reduction is consistent with the learned strengthening of cause–effect structures, indicating a shift toward more predictable token transitions.

\textbf{Finally,} the TPM construction policy and the choice of reduced dimensionality are shown to have a significant impact on evaluation outcomes. While the Trajectory mode leads to a pronounced reduction in response length, the Batch mode preserves model accuracy more effectively. Moreover, increasing the dimensionality of the TPM substantially raises sample complexity requirements, underscoring the importance of selecting an appropriate dimensionality that is well matched to the amount of data generated in a single training step.


\begin{figure*}[t]
\centering
\fbox{%
\begin{minipage}{0.96\textwidth}
\small

\textbf{Prompt.} \emph{<Cecilia just bought a new puppy. According to her veterinarian, she has to feed the puppy 1 cup of dog food every day for the first 180 days. Then she has to feed the puppy 2 cups of dog food every day for the rest of its life. If one bag of dog food contains 110 cups, how many bags of dog food will Cecilia use in the first year? Let's think step by step and output the final answer after "\#\#\#\#".>}\\[0.6em]

\textbf{Baseline} \hfill \textbf{Tokens: 326} \hfill \textbf{Label: \textcolor{red}{Incorrect}}\\
\begingroup\ttfamily
\noindent <Step 1: Calculate the number of days in the first 180 days.

There are 365 days in a year. So, 365 days - 180 days = 185 days in the first year.

Step 2: Calculate the number of cups Cecilia needs for the first 180 days.

The puppy needs 1 cup of dog food every day for the first 180 days. So, 1 cup/day * 180 days = 180 cups.

Step 3: Calculate the number of days left in the year.

There are 185 days - 180 days = 5 days left in the year.

Step 4: Calculate the number of cups Cecilia needs for the remaining 5 days.

The puppy needs 2 cups of dog food every day for the remaining 5 days. So, 2 cups/day * 5 days = 10 cups.

Step 5: Calculate the total number of cups Cecilia will need in the first year.

There are 180 cups for the first 180 days + 10 cups for the remaining 5 days = 190 cups.

Step 6: Calculate the number of bags Cecilia will use.

One bag of dog food contains 110 cups. So, 190 cups / 110 cups/bag = 1.73 bags.

Since Cecilia cannot use a fraction of a bag, she will need 2 bags of dog food in the first year.

Final Answer: Cecilia will use 2 bags of dog food in the first year. \#\#\#\# 2>
\par\endgroup
\vspace{0.8em}

\textbf{Ours} \hfill \textbf{Tokens: 117} \hfill \textbf{Label: \textcolor{green!50!black}{Correct}}\\
\begingroup\ttfamily
\noindent <Total number of days in a year = 365 days

First 180 days => 180 days * 1 cup/day = 180 cups

Rest of the year = 365 days - 180 days = 185 days

Rest of the year => 185 days * 2 cups/day = 370 cups

Total amount of dog food in first 180 days + rest of the year = 180 + 370 = 550 cups

One bag contains 110 cups, so number of bags = 550 / 110 = 5 bags

\#\#\#\# 5>
\par\endgroup

\end{minipage}}
\caption{Example showing fewer generated tokens and corrected final answer compared to the baseline.}
\label{fig:gsml_sample}
\end{figure*}

\end{document}